\documentclass{sig-alternate}
\usepackage[numbers,sort&compress]{natbib}
\usepackage{graphicx}
\usepackage{subfigure}
\usepackage{enumitem}
\usepackage{color}
\usepackage[linesnumbered,ruled]{algorithm2e}
\usepackage{ amssymb }
\usepackage{ mathrsfs }
\usepackage{float}

\usepackage{fancyhdr}
 
\begin{document}

\title{Hardware-Software Codesign of Accurate, Multiplier-free Deep Neural Networks}

\numberofauthors{4}
\author{
\alignauthor
Hokchhay Tann\\
       \affaddr{School of Engineering}\\
	   \affaddr{Brown University}\\
       \affaddr{Providence, RI}\\
       \email{hokchhay\_tann@brown.edu}
\alignauthor
Soheil Hashemi\\
       \affaddr{School of Engineering}\\
       \affaddr{Brown University}\\
       \affaddr{Providence, RI}\\
       \email{soheil\_hashemi@brown.edu}
\and
\alignauthor
R. Iris Bahar\\
       \affaddr{School of Engineering}\\
       \affaddr{Brown University}\\
       \affaddr{Providence, RI}\\
       \email{iris\_bahar@brown.edu}
\alignauthor Sherief Reda\\
       \affaddr{School of Engineering}\\
       \affaddr{Brown University}\\
       \affaddr{Providence, RI}\\
       \email{sherief\_reda@brown.edu}
}
\author{Hokchhay Tann, Soheil Hashemi, R. Iris Bahar, Sherief Reda 
\\
      \affaddr{School of Engineering,}
      \affaddr{Brown University,}
      \affaddr{Providence RI 02912}\\
      \email{\large\{hokchhay\_tann, soheil\_hashemi, iris\_bahar, sherief\_reda\}@brown.edu}
}

\maketitle

\begin{abstract}
While Deep Neural Networks (DNNs) push the state-of-the-art in many machine learning applications, they often require millions of expensive floating-point operations for each input classification. This computation overhead limits the applicability of DNNs to low-power, embedded platforms and incurs high cost in data centers. This motivates recent interests in designing low-power, low-latency DNNs based on fixed-point, ternary, or even binary data precision. While recent works in this area offer promising results, they often lead to large accuracy drops when compared to the floating-point networks. We propose a novel approach to map floating-point based DNNs to 8-bit dynamic fixed-point networks with integer power-of-two weights with no change in network architecture. Our dynamic fixed-point DNNs allow different radix points between layers. During inference, power-of-two weights allow multiplications to be replaced with arithmetic shifts, while the 8-bit fixed-point representation simplifies both the buffer and adder design. In addition, we propose a hardware accelerator design to achieve low-power, low-latency inference with insignificant degradation in accuracy. Using our custom accelerator design with the CIFAR-10 and ImageNet datasets, we show that our method achieves significant power and energy savings while increasing the classification accuracy.
\end{abstract}

\section{Introduction}
\label{sec:Introduction}
Recent availability of high-performance computing platforms has enabled the success of deep neural networks (DNNs) in many demanding fields, especially in the domains of machine learning and computer vision. At the same time, applications of DNNs have proliferated to platforms ranging from data centers to embedded systems, which open up new challenges in low-power, low-latency implementations that can maintain state-of-the-art accuracy. While systems with general purpose CPUs and GPUs are capable of processing very large DNNs, they have high power requirements and are not suitable for embedded systems, which has led to increasing interest in the design of low-power custom hardware accelerators.

In designing low-power hardware for DNNs, one major challenge stems from the high precision used in the network parameters. State-of-the-art DNNs in classification accuracy are typically implemented using single precision (32-bit) floating-point, which requires large memory size for both the network parameters as well as the intermediate computations. Complex hardware multipliers and adders are also needed to operate on such representations. 

On the other hand, the inherent resiliency of DNNs to insignificant errors, has resulted in a wide array of hardware-software codesign techniques targeted for lowering the energy and memory footprint of these networks. Such techniques broadly aim either to lower the cost of each operation by reducing the precision \cite{Gysel,courbariaux,hashemi2017} or to lower the number of required operations, for example by knowledge distillation \cite{bucilua2006model,hinton2015distilling,fitnet}.

While previous studies offer low-precision DNNs with little reduction in accuracy, the smallest fixed-point solutions proposed require 8-bits or more for both the activation and network parameters. Furthermore, while solutions with binary and ternary precisions prove effective for smaller networks with small datasets, they often lead to unacceptable accuracy loss on large datasets such as ImageNet \cite{ILSVRC15}. In addition, these low-precision network techniques usually require precision specific network designs and therefore cannot readily be used on a specific network without an expensive architecture exploration. 

In this work, we aim to tackle the low-power high-accuracy challenge for DNNs by proposing a hardware-software codesign solution to transform existing floating-point networks to 8-bit dynamic fixed-point networks with integer power-of-two weights without changing the network topology. The use of power-of-two weights enables a multiplier-free hardware accelerator design, which efficiently performs computation on dynamic fixed-point precision. More specifically, our contributions in this paper are as follows:
\begin{itemize}[noitemsep]
\item We propose to compress floating-point networks to 8-bit dynamic fixed-point precision with integer power-of-two weights. We then propose to fine-tune the quantized network using student-teacher learning to improve classification accuracy. Our technique requires no change to the network architecture.
\item We propose a new multiplier-free hardware accelerator for DNNs and synthesize it using an industry level library. Our custom accelerator efficiently operates using 8-bit multiplier-free dynamic fixed-point precision. 
\item We also propose to utilize an ensemble of dynamic fixed-point networks, resulting in improvements in classification accuracy compared to the floating-point counterpart, while still allowing large energy savings.
\item We evaluate our methodologies on two state-of-the-art and demanding test sets, namely CIFAR-10 and ImageNet, and use well-recognized network architectures for our experiments. We compare our solution against a baseline floating-point accelerator and quantify the power and energy benefits of our methodology.
\end{itemize}

The rest of our paper is organized as follows. In Section \ref{sec:Background}, we provide a breif background on deep neural networks. In Section \ref{sec:related_work}, we summarize previous work related to ours. Section \ref{sec:methodology} describes our methodologies and accelerator design. Next, in Section \ref{sec:results}, we provide the results obtained from our methodologies and our custom accelerator. Here we discuss the performance from both hardware and accuracy perspectives. Finally, in Section \ref{sec:conc} we conclude our work.

\section{Background}
\label{sec:Background}
\begin{figure}[t]
   \begin{center}
    \includegraphics[scale=0.25]{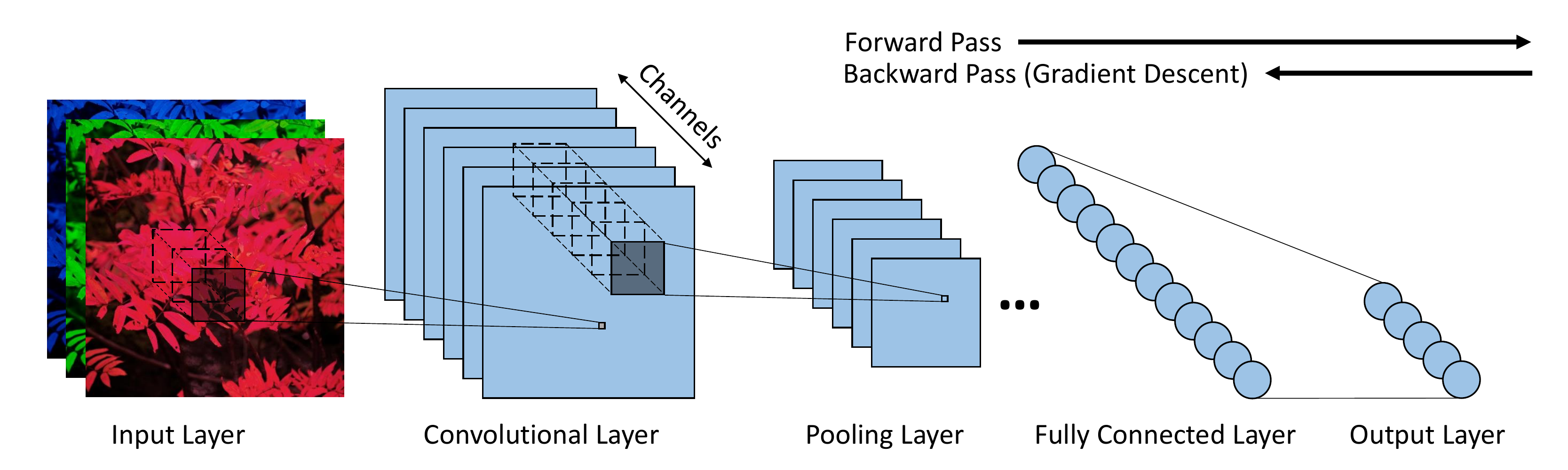}
    \caption{Typical Structure of Deep Neural Network}
    \label{fig:dnn}
    \end{center}
    \vspace{-0.3in}
\end{figure}

Figure \ref{fig:dnn} shows the template structure of a deep neural network. While a large number of layer types are available in the literature, three layer types are more commonly used in DNNs:

\begin{itemize}[noitemsep]
\item\textit{Convolutional Layers}: Each neuron in this layer is connected to a subset of inputs with the same spatial dimensions as the kernels, which are typically 3-dimensional as shown in  Figure \ref{fig:dnn}. The convolution operation can be formulated as $y=b+\sum_{i}\sum_{j}\sum_{k}(\textbf{x}_{i,j,k}\cdot\textbf{w}_{i,j,k})$. Here, $\textbf{x}$ is the input subset, $\textbf{w}$ is the kernel weight matrix, and $b$ is a scalar bias. These layers are used for feature extractions.

\item\textit{Pooling Layers}: Pooling layers are simply used to down sample input data.

\item\textit{Fully-Connected Layers}: These layers are similar to convolutional layers with differences being that inputs and kernels are one-dimensional vectors. These layers are often used toward the end as classifier, where the output vector from the final layer (logits) is fed to a logistic function.

\item\textit{Non-Linearity}: For each scalar input $x$, this layer outputs $\sigma(x)$, where $\sigma(\cdot)$ is a predefined non-linear function, such as $tanh(\cdot)$, rectify linear unit (ReLU), etc.
\end{itemize}


DNNs typically are based on floating-point precision and trained with backpropagation algorithm. Each training step involves two phases: forward and backward. In the forward phase, the network is used to perform classification on the input. Afterward, the gradients are propagated back to each layer in the backward phase to update the network's parameters. The biggest portion of the computational demands are required by the multiplier blocks utilized in the convolutional and fully connected layers.

\section{Related Works}
\label{sec:related_work}
Previous work in software and hardware implementation of DNNs has been, for the most part, disconnected. Few studies have tried to optimize highly accurate designs with low power budgets. On the accuracy front, one aspect of condensing DNNs is to train much smaller networks from the large, cumbersome models \cite{bucilua2006model,hinton2015distilling}. Both models are based on floating-point precision. This approach proposes to train the student (smaller model) to mimic to the outputs of the teacher (larger model). The loss function for the training is composed of two parts: the losses with respect to the true labels and the outputs from the teacher model.

Alternatively, DNNs with low precision data formats have enormous potentials for reducing hardware complexity, power and latency. Not surprisingly, there exists a rich body of literature which studies such limited precisions. 
Previous work in this area have considered a wide range of reduced precision including fixed point~\cite{courbariaux2014low,Gupta,tang1993multilayer}, ternary (-1,0,1) \cite{hwang2014fixed} and binary (-1,1) \cite{courbariaux,soudry2014expectation}. Furthermore, comprehensive studies of the effects of different precision on deep neural networks are also available. Gysel \textit{et al.} \cite{Gysel} propose Ristretto, a hardware-oriented tool capable of simulating a wide range of signal precisions. While they consider dynamic fixed-point, in their work the focus is on network accuracy so the hardware metrics are not evaluated. On the other hand, Hashemi \textit{et al.} \cite{hashemi2017} provide a broad evaluation of different precisions and quantizations on both the hardware metrics and network accuracy. However, they do not evaluate dynamic fixed point.

In the hardware design domain, while few work have considered different bit-width fixed-point representations in their accelerator design \cite{hashemi2017,Sankaradas:2009,Zhang:2015}, in contrast to the accuracy analysis, no evaluation of hardware designs using dynamic fixed-point is available. We fill this gap by providing an accelerator design optimized to use dynamic fixed-point representation for intermediate computations while using power-of-two weights.

In recent years, a few works have focused on techniques to reduce the power demands of DNNs at the cost of small reductions in network accuracy. For instance, Tann {\it et al.} propose an incremental learning algorithm where the network in trained in incremental steps~\cite{TANN:16}. The idea is then to turn off large portions of the network in order to save energy if these portions are not needed to retain accuracy.  
While this work delivers significant power and energy saving with small network accuracy degradation, it is orthogonal to our work and can be applied in conjunction. 

Sarwar {\it et al.} propose a multiplier-less neural network where an accurate multiplier is replaced with an alphabet set multiplier to save power. This work however, focuses on multi-layer perceptrons and deep neural networks are not evaluated~\cite{Sarwar:2016}. In contrast, we evaluate our work on both CIFAR-10 and ImageNet and highlight that our methodology is capable of delivering significant savings in energy while even showing improvements in accuracy.

\section{Multiplier-Free Dynamic Fixed-Point (MF-DFP) Networks}
\label{sec:methodology}
In order to simplify the hardware implementation, we propose to alter the compute model by replacing multipliers with shift blocks and reducing signal bit width to 8 bits. We represent the signals using dynamic fixed-point format since synaptic weights and signals in different layers can vary greatly in range. Employing a uniform fixed-point representation across the layers would require large bit widths to accommodate for such range. As demonstrated by others~\cite{Gysel, hashemi2017}, even with 16-bit fixed-point, significant accuracy drop is observed when compared to floating-point representation.

Dynamic fixed-point representation, as proposed in \cite{courbariaux2014low}, can be represented using two variables $\langle b, f\rangle$, where $b\in \mathbb{N+}$ is the bit width, and $f \in \mathbb{Z}$ is the fractional length. Each $b$-bit number in this scheme can be interpreted as $(-1)^s2^{-f}\sum_{i=0}^{b-2}2^{i}x_{i}$, where $s,x_i\in \{0,1\}$ are the sign bit and the $i^{th}$ bit respectively. The term dynamic refers to the fact that different layers in DNNs can take on different values for $f$ depending on their ranges. 
In this work, we deploy 8-bit dynamic fixed-point for all of our experiments.

While we adopt our quantization process from the techniques in \cite{Gysel}, our work differs from theirs in three aspects: ($i$) we perform hardware-software analysis for power-of-two weights and dynamic fixed-point data path, ($ii$) we propose to include student-teacher learning in the fine-tuning process, and ($iii$) we demonstrate that an ensemble of two MF-DFP networks can outperform the floating-point network while having significant savings in energy. These aspects are described in Algorithm \ref{alg:construction} as three phases. Next, we describe these phases in more details.

\subsection{Network Quantization (Phase 1)} \label{ssec:construction}

In order to construct a dynamic fixed-point network, we take as input a fully trained floating-point network. We first quantize on this input network by rounding its weights to the nearest powers of two. We also round the intermediate signals to 8-bit dynamic fixed-point using Ristretto~\cite{Gysel} (line \ref{alg:quantize}). We then perform fine-tuning on the network to recover from accuracy loss due to quantization (lines \ref{alg:phase1}--\ref{alg:endphase1}).

DNNs are typically trained using the backpropagation algorithm with variants of gradient descent methods, which can be ill-suited for low-precision networks. The computed gradients and learning rates are typically very small, which means that parameters may not be updated at all due to their low-precision format. Intuitively, this requires high precision in order to converge to a good minima. However, integer power-of-two weights only allow large increment jumps.

To combat this disparity, we adopt solution proposed by Courbariaux \textit{et al.}~\cite{courbariaux} to keep two sets of weights during the training process: one in quantized precision and one in floating-point. As shown in Algorithm \ref{alg:construction}, during forward propagation, the floating-point weight set is stochastically or deterministically quantized before the input data is evaluated (line \ref{alg:forward}). For our work, we found that deterministic quantization gives better performance. The output result of the quantized network is then used to compute the loss with respect to the true label of the data (line \ref{alg:grad}). The gradients with respect to this loss are then used to update the floating-point parameters during backward propagation (line \ref{alg:backward}), and the process is repeated until convergence. This approach allows small gradients to accumulate over time and eventually cause incremental updates in the quantized weights.

\subsection{Additional Fine-tuning (Phase 2)} \label{ssec:finetune}

On top of the technique from Courbariaux \textit{et al.}~\cite{courbariaux}, we propose additional training with a different loss function once training with hard labels no longer improves the performance. As shown in Algorithm~\ref{alg:construction} lines \ref{alg:phase2}--\ref{alg:endphase2}, in addition to using hard labels, we introduce student-teacher learning, where a \textit{student} network is trained to mimic the outputs of a \textit{teacher} network \cite{bucilua2006model, hinton2015distilling}. Both networks are floating-point based, but the \textit{student} typically has a far fewer number of parameters. In our work, we treat the dynamic fixed-point network as the \textit{student} and the floating-point network as the \textit{teacher}.

\setlength{\textfloatsep}{0pt}
\begin{algorithm}[t]
\label{alg:construction}
    \SetKwBlock{Begin}{begin}{end}
    \SetKwInOut{Input}{Input}
    \SetKwInOut{Output}{Output}
    \tcp{FLnet: the input floating-point net.}
    \tcp{t\_logits: the floating-point net's logits.}
    \Input{FLnet, t\_logits}\label{alg:input}
    \Begin(Phase 1){\label{alg:phase1}
    \tcp{MF\_DFPnet: dynamic fixed-point network}
    \Output{MF\_DFPnet}   
    \tcp{Quantize the weights to powers of two and}
    \tcp{inputs to 8-bit fixed-point}

    MF\_DFPnet = Quantize\_8bit(FLnet); \label{alg:quantize}\\

    \tcp{Fine-tuning MF\_DFPnet until convergence}
    \tcp{using hard data labels as in \cite{courbariaux}.}
    \For{i = 1 to Convergence} 
      {\label{alg:for1}
      Forward\_Pass(MF\_DFPnet);\label{alg:forward}\\
      \tcp{Compute gradients}
      grads=Grad(MF\_DFPnet, true\_labels);\label{alg:grad}\\
      \tcp{Backpropagate grad and update weights:}
      Backward\_update(FLnet, grads);\label{alg:backward}\\
      MF\_DFPnet = Quantize\_8bit(FLnet);\\
      }
    } \label{alg:endphase1}
    \Begin(Phase 2){\label{alg:phase2}
    \tcp{Additional training using different}
    \For{j = i to Convergence}
      {\label{alg:for2}
      Forward\_Pass(MF\_DFPnet);\\
      grads=Grad(MF\_DFPnet, true\_labels);\\
      \tcp{Also with respect to teacher's logits}
      grad\_logit=Grad(MF\_DFPnet, t\_logits);\\
      \tcp{gradient to update:}
      gradients=grads + $\beta\cdot$grad\_logit;\\
      Backward\_update(FLnet, gradients);\\
      MF\_DFPnet = Quantize\_8bit(FLnet);\\
      } \label{alg:endfor2}
    \Return MF\_DFPnet \label{alg:end}\\
    }\label{alg:endphase2}
    \tcp{\textbf{Phase 3:} If using ensemble, repeat Phase 1}
    \tcp{and 2 with different input FLnet.}
    \caption{Floating-Point to Dynamic Fixed-Point}
\end{algorithm}

The loss function in the student-teacher learning incorporates the knowledge learned by the \textit{teacher} model \cite{hinton2015distilling}. Suppose S is the \textit{student} network, and T is the \textit{teacher} with output logit vectors $z_S$ and $z_T$ and class probability $P_S$ and $P_T$ respectively. The softmax regression function is relaxed by introducing a temperature parameter $\tau$ such that $P_{S,i}=\frac{exp(z_{S,i}/\tau)}{\sum_{j}exp(z_{S,j}/\tau)}$ and $P_{T,i}=\frac{exp(z_{T,i}/\tau)}{\sum_{j}exp(z_{T,j}/\tau)}$. Let $\textbf{W}_S$ be the parameters of the student network, then the loss function for the student model is defined to be:
\begin{align} \label{eq:newloss}
\mathcal{L}(\textbf{W}_S)=\mathcal{H}(\mathbf{Y},P_S) + \beta\cdot\mathcal{H}(P_T, P_S)
\end{align}
where $\beta$ is a tunable parameter, $\mathcal{H}$ is the cross entropy, and $\mathbf{Y}$ is the one-hot true data label. Using $\tau >> z_S, z_T$, we have $P_i=\frac{exp(z_{i}/\tau)}{\sum_{j}exp(z_{j}/\tau)}\approx\frac{1+z_{i}/\tau}{N+\sum_{j}z_{j}/\tau}$ where $N$ is the length of of vectors $z_S, z_T$. With zero-meaned $z_S, z_T$ ($\sum_jz_{S,j}=\sum_{j}z_{T,j}=0$), the approximated gradient is then:
\begin{align}
\frac{\delta\mathcal{L}(\textbf{W}_S)}{\delta z_{S,i}}\approx(P_{S,i}-\mathbf{Y}_i) + \frac{\beta}{N\cdot \tau^2}\cdot(z_{S,i} - z_{T,i}).
\end{align}

\subsection{Ensemble of MF-DFP Networks (Phase 3)} \label{ssec:ensemble}
Deploying an ensemble of DNNs has been proven to be a simple and effective method to boost the inference accuracy of a DNN \cite{deepdnn?}. The idea is to independently train multiple DNNs of the same architecture and use them to evaluate each input. The output is then chosen based on the majority of votes. Suppose the ensemble consists of $M$ networks producing output logit vectors $\textbf{z}_i$, $i\in[1,M]$. Then the output class can simply be the maximum element in $\frac{1}{M}\sum_{i=1}^M\textbf{z}_i$. 

This idea is amenable in scenarios where there exists enough time or energy budget to justify evaluating the input on a number of networks. In section \ref{ssec:results} we highlight that, since the reduction in energy from the proposed MF-DFP are so dramatic,  the designer may implement an ensemble of MF-DFP networks in parallel and still save significantly in energy consumption.  More specifically, we show that an ensemble of multiplier-free dynamic fixed-point networks can outperform a floating-point network while still achieving significant energy saving. In order to construct such ensemble, we run Algorithm \ref{alg:construction} multiple times with different starting floating-point networks on line \ref{alg:input}.

\section{Hardware Accelerator Design}
\label{ssec:hardware}
As discussed in Section \ref{sec:methodology}, while we maintain low-precision in both network signals and parameters for efficiency, providing the network with the flexibility to change the location of the radix point from layer to layer is necessary for minimizing the accuracy degradation. While improving the accuracy, this scheme incurs complexities in the hardware design as some bookkeeping in needed to keep track of the location of the radix point in different parts of the network. In the proposed accelerator, we enable such flexibility by providing each set of calculations with details on the indices of both the input feature maps as well as the output activation. More specifically, we implement this feature by adding control signals dedicated to both the input feature, and the output activation radix indices. Dedicated hardware is then added to the hardware to shift the result to the correct index as determined by the radix indices.

On the other hand, while dynamic fixed-point representation for synaptic weights and activation maps allows for compact bit widths, during inference, we would still need to perform fixed-point multiplications. As described in Section~\ref{sec:methodology}, we propose to quantize the weights to integer power-of-two, which would allow the expensive multiplications to be replaced with arithmetic shifts. These shift operators are far more hardware-friendly than full-scale multipliers. In this quantization scheme, for each weight $w$, we represent its quantized version using two numbers $\langle s,e\rangle$, where $s$ is the sign of the weight $w$, and $e=max[round(\log_2(|w|)),-7]$ is the exponent for the power of 2 (i.e., $2^e$). Here, $round()$ performs rounding to the nearest integer. Note that we bound $e\ge -7$ since our input data is limited to 8 bits. For each input $x$, $x\cdot w$ is then transformed into $(s\cdot x)\ll \gg e$, where $\ll \gg$ represents the shift operator. In addition, we observe that the magnitudes of the weights is less than 1, so our rounding leads to 8 possible exponents $\{0,-1,\ldots,-7\}$. Therefore the weights can be encoded into 4-bit representation. This observation is used to simplify our hardware architecture significantly as discussed in \ref{ssec:results}.

\begin{figure}[t]
   \begin{center}
    \includegraphics[scale=0.25]{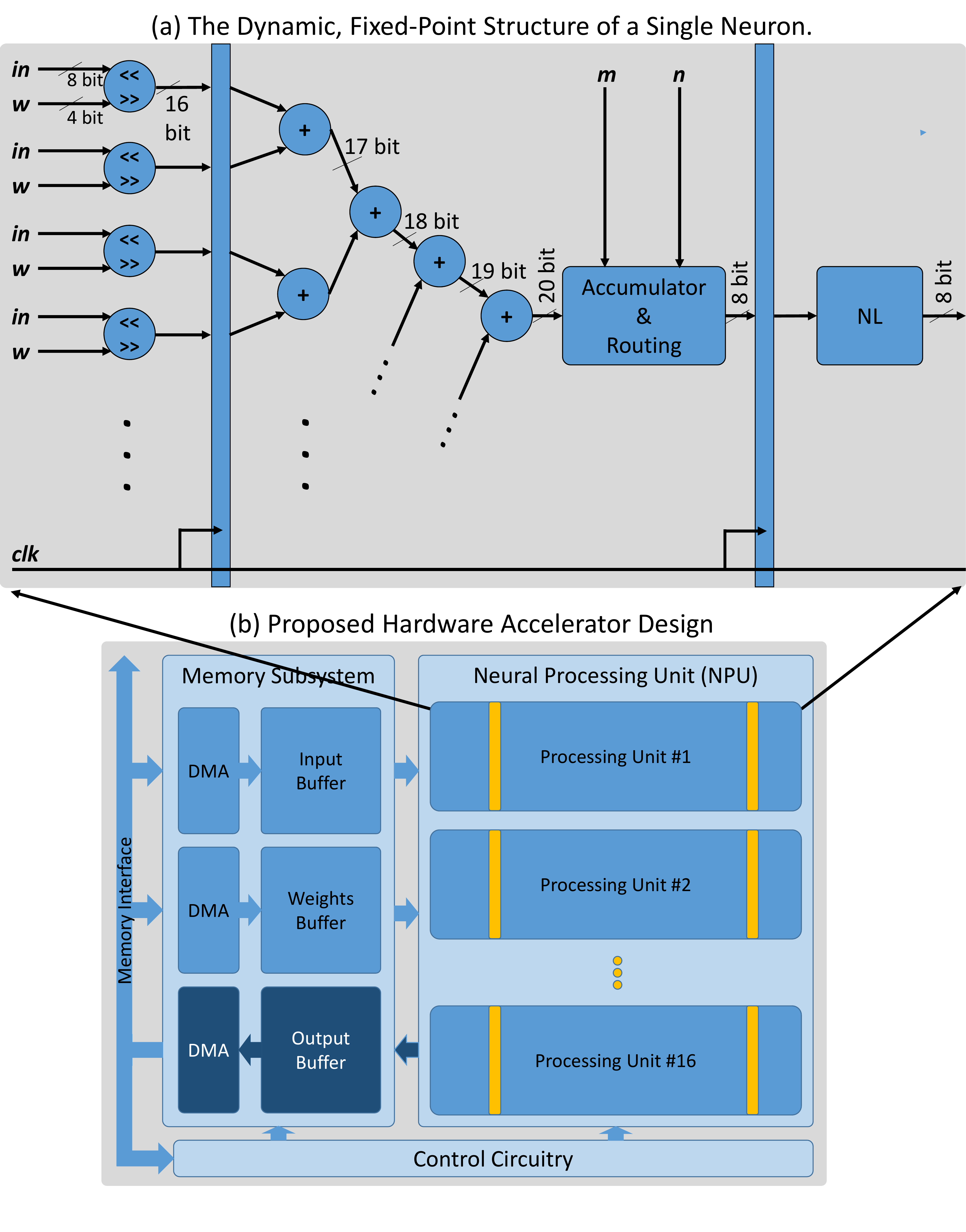}
    \vspace{-0.3in}
    \caption{Proposed Hardware Accelerator Design (a) a Single Neuron and (b) The Organization of Neurons and Hardware Blocks.}
    \label{fig:hw}
    \end{center}
\end{figure}

To further improve the accuracy, we ensure that there is no loss in intermediate values by mitigating the possibility of overflows. In order to do so, we ensure that all intermediate signals have large enough word-width, thereby effectively increasing the width of the intermediate wires as needed. To illustrate our idea, Figure~\ref{fig:hw}(a) shows the simplified structure of a single neuron in our proposed implementation, highlighting the main feature of the accelerator design. In Figure~\ref{fig:hw}(a), the dedicated hardware implementing the dynamic fixed-point scheme is shown as ``Accumulator \& Routing''. Here $m$ and $n$ represent the locations of the radix points for the input features and output activations respectively. 

In order to integrate our proposed neuron architecture into a full-scale hardware accelerator, we utilize a tile-based implementation inspired from DianNao \cite{diannao}, where each cycle a small number of physical neurons is fed a new set of data for calculation. We implement three separate memory subsystems assigned to input data, weights, and output data, respectively. This memory subsystem ensures the isolation of memory transfers from the calculation for maximum throughput. The computation itself is performed in neural processing units (NPUs) containing a number of processing units each implementing 16 neurons with 16 synapses. 

Figure~\ref{fig:hw}(b) illustrates the organization of the proposed hardware accelerators. Here we want to stress the benefits of our methodologies relative to the floating-point design. Thus, an architectural design space exploration such as altering number of hardware neurons and synapses is out of the scope of this work.

In order to incorporate the proposed ensemble of networks, the number of processing units is increased as needed to parallelize the computation of an ensemble of networks. Note that the memory subsystems as well as the control logic also need to be modified to account for the number of processing units. In section \ref{ssec:results} we evaluate our methodologies using a single processing unit, resulting in a single multiplier-free dynamic fixed point (MF-DFP) network, and two processing units, which form an ensemble of two networks.

We also implement and compare our hardware design with a conventional 32-bit floating-point architecture using a single processing unit as a baseline. Compared to our proposed design, the baseline implementation utilizes multipliers in the first stage of the design and keeps the bitwidth constant at 32-bits throughout the design for both the activations and the network parameters.

\section{Experimental Results}
\label{sec:results}

\subsection{Experimental Setup}
\label{ssec:setup}
In this section, we discuss our results on the CIFAR-10 and ImageNet 2012 datasets \cite{cifar10, ILSVRC15} using the well-known DNN architectures from \cite{cifar10} and \cite{Krizhevsky} respectively. We remove all local response normalization layers since they are not amenable to our multiplier-free hardware implementation. All of our experiments are based on Caffe \cite{caffe}.

For CIFAR-10, we begin by training the floating-point networks using the benchmark architecture. For the ImageNet benchmark, we obtain the floating-point model from Caffe Model Zoo\footnote{https://github.com/BVLC/caffe/wiki/Model-Zoo}. We then run the networks on their corresponding training set data to obtain the pre-softmax output logits. From these floating-point networks, we construct our proposed MF-DFP networks using Algorithm \ref{alg:construction}.

For our hardware evaluations, we compile our designs using Synopsys Design Compiler and a 65 nm standard cell library in the typical processing corner. We synthesize our hardware so that we have zero timing slack for the floating-point design. Therefore, we use a constant clock frequency of 250MHz for all our experiments. While the utilization of barrel shifters instead of multipliers provides us with timing slacks which can be used to boost the frequency, we choose to keep the frequency constant as changing the frequency adds another dimension for evaluation which is out of the scope of this work.

\subsection{Results}
\label{ssec:results}
We evaluate our proposed methodology as well as our custom hardware accelerator on CIFAR-10 and ImageNet using a broad range of performance metrics including accuracy, power consumption, design area, inference time, and inference accuracy. Table \ref{table:power} summarizes the design area and the power consumption of the proposed multiplier-free custom accelerator. Values shown in parenthesis, (in,w), reflect the number of bits required for the representation of inputs and weights respectively. We also implement a floating-point version of our accelerator as a baseline design and for comparison. As shown in the table our accelerator can achieve significant benefits in both design area and power consumption using both one processing unit and using an ensemble of two networks. Next we report the results when using our methodologies and hardware accelerator for our benchmarks.

\begin{table}[t]
  \small
  \caption{Design metrics of the proposed MF-DFP accelerator against the floating-point baseline.}
  \centering
  \begin{tabular}{l|r|r|r|r}
    \hline
     	  & \multicolumn{1}{|l|}{Design} & \multicolumn{1}{|l|}{Power} & \multicolumn{1}{|l|}{Area} & \multicolumn{1}{|l}{Power} \\
		 & \multicolumn{1}{|l|}{Area} & \multicolumn{1}{|l|}{Cons.} & \multicolumn{1}{|l|}{Saving} & \multicolumn{1}{|l}{Saving} \\
Precision ($in,w$) & \multicolumn{1}{|l|}{($mm^2$)} & \multicolumn{1}{|l|}{($mW$)} & \multicolumn{1}{|l|}{($\%$)} & \multicolumn{1}{|l}{($\%$)} \\
    \hline
    Floating-point(32,32)  & 16.52 & 1361.61 & 0 & 0 \\
    Proposed MF-DFP(8,4)   & 1.99 & 138.96 & 87.97 & 89.79 \\
    Ens. MF-DFP(8,4) & 3.96 & 270.27 & 76.00 & 80.15 \\
    \hline
  \end{tabular}
  \label{table:power}
  \vspace{-0.1in}
\end{table}

\begin{table*}[t]
  \small
  \caption{Time, energy and top-1 accuracy for CIFAR-10 and ImageNet. In addition, for Imagenet, we also show in parenthesis the top-5 accuracy. Ensemble accuracy is obtained by deploying two MF-DFP networks.}
  \centering
    \begin{tabular}{l|r|r|r|r|r|r|r|r}
    \hline
     	  & \multicolumn{4}{|c|}{CIFAR-10} & \multicolumn{4}{|c}{ImageNet} \\ \cline{2-9}
 & \multicolumn{1}{|l|}{Classification} & \multicolumn{1}{|l|}{Time} & \multicolumn{1}{|l|}{Energy} & \multicolumn{1}{|l|}{Energy} & \multicolumn{1}{|l|}{Classification} & \multicolumn{1}{|l|}{Time} & \multicolumn{1}{|l|}{Energy} & \multicolumn{1}{|l}{Energy} \\
Precision & \multicolumn{1}{|l|}{Accuracy ($\%$)} & \multicolumn{1}{|l|}{($us$)} & \multicolumn{1}{|l|}{($uJ$)} & \multicolumn{1}{|l|}{Saving ($\%$)} & \multicolumn{1}{|l|}{Accuracy ($\%$)} & \multicolumn{1}{|l|}{($us$)} & \multicolumn{1}{|l|}{($uJ$)} & \multicolumn{1}{|l}{Saving ($\%$)} \\
    \hline
    Floating-Point (32,32) & 81.53 & 246.52 & 335.68 & 0    & 56.95 (79.88) & 15666.45 & 21332.38 & 0 \\
    MF-DFP (8,4)           & 80.77 & 246.27 & 34.22 & 89.81 & 56.16 (79.13) & 15666.06 & 2176.96 & 89.80 \\
    Ensemble MF-DFP        & 82.61 & 246.27 & 66.56 & 80.17 & 57.57 (80.29) & 15666.06 & 4234.07 & 80.15 \\
    \hline
  \end{tabular}
  \vspace{-0.2in}
  \label{table:results}
\end{table*}

\begin{figure}[t]
   \begin{center}
    \includegraphics[scale=0.59,trim=16 0 9 17,clip]{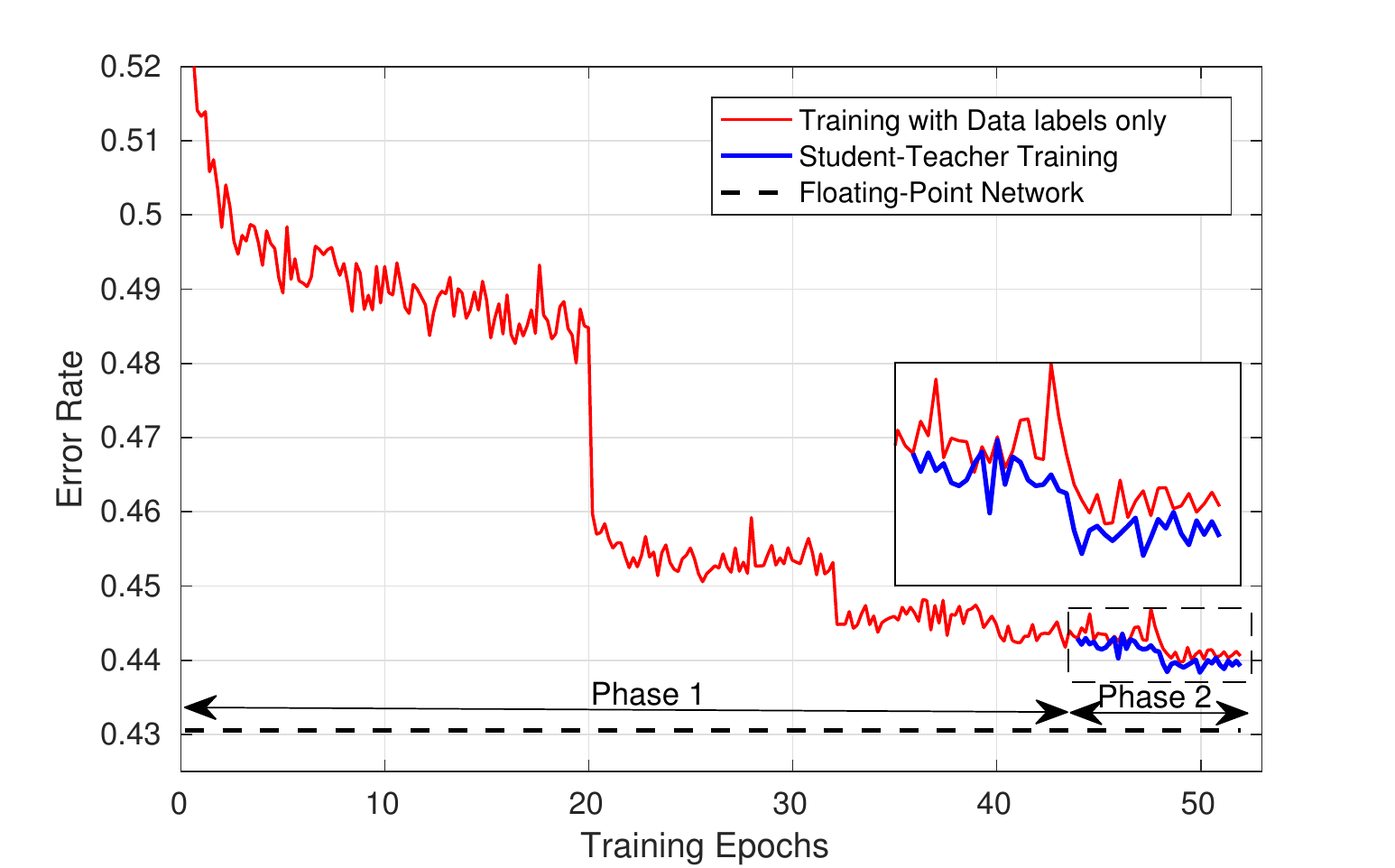}
     \vspace{-0.21in}
    \caption{ImageNet 2012 validation set top-1 error rates for quantized network trained with only data labels, with student-teacher learning, and floating-point network.}
    \label{fig:training}
    \end{center}
    \vspace{-0.05in}
\end{figure}

Figure \ref{fig:training} shows the classification error rate of the baseline floating-point network as well as the fine-tuning process of MF-DFP for the ImageNet benchmark. Here, we observe that by fine-tuning using just data labels (Phase 1), we achieve significant performance with less than a 1\% increase in error rate than the floating-point counterpart. Additional training using the student-teacher model (Phase 2) as described in Section \ref{ssec:finetune} on top of just data labels, allows us to reduce the error rate even more. In this experiment, we observed that more benefit is achieved when the student-teacher training is started from a non-global optimal point in the data labels-only training. More specifically, the value of $i$ in Algorithm \ref{alg:construction} line \ref{alg:for2} should be close to convergence but not the global optimal point in the training process. In either case, the student-teacher learning provides consistently better performance than using the data labels-only training. For this training, we set $\tau=20$, $\beta=0.2$ and start with a learning rate of 1e-03. We decrease the rate by a factor of 10 when learning levels off and stop the training when the learning rate drops below 1e-07.

Furthermore, in Table \ref{table:results}, we summarize the accuracy, inference time, and the energy performance of our proposed techniques. As shown in the table, our methodology can achieve energy savings as high as 89\% in the case of single MF-DFP network with a maximum of 0.79\% degradation in accuracy for both benchmarks. This is especially significant as there is absolutely no modification to network depth and channel size. In addition, with the extra area budget, we can implement two processing units in our accelerator and, for each benchmark, we deploy an ensemble of two MF-DFP networks trained using different starting points. As shown in Table \ref{table:results}, we can outperform the floating networks in both benchmarks using this ensemble while still achieving significant energy saving.



Finally, while we design our methodology with memory footprint in mind, we do not include the power consumption of the main memory subsystem in our evaluations. However, as a general guideline, our methodology emphasizes on reductions in network precisions and therefore requires 8$\times$ less memory compared to a floating-point implementation as shown in Table \ref{table:memory}. For the ensemble method, the memory requirement essentially doubles from single MF-DFP, however, they are still far lower than the floating-point networks.

\begin{table}[t]
  \small
  \caption{Comparison of memory requirements for floating-point versus MF-DFP network parameters for CIFAR-10 and ImageNet benchmarks.}
   \centering
    \begin{tabular}{l|r|r}
    \hline
Precision & \multicolumn{1}{|l|}{CIFAR-10 (MB)} & \multicolumn{1}{|l}{ImageNet (MB)} \\
    \hline
    Floating-Point  & 0.3417 & 237.95  \\
    MF-DFP            & 0.0428 & 29.75 \\
    Ensemble MF-DFP   & 0.0855 & 59.50 \\
    \hline
   \end{tabular}
   \vspace{0.05in}
  \label{table:memory}
\end{table}

\section{Conclusion}
\label{sec:conc}
In this work we proposed a novel hardware-software codesign approach which enables seamless mapping of full-precision deep neural networks to a multiplier-free dynamic fixed-point network. In our work, no change to the network architecture is required to maintain accuracy within acceptable bounds. We also formalized the use of student-teacher learning for accuracy improvements in low-precision networks. In addition, we proposed a hardware design capable of incorporating the dynamic fixed point as well as the multiplier-free design aspects. We proposed to utilize an ensemble of lower precision ML-DFP networks to increase the accuracy even further. We evaluated our designs using two well-recognized and demanding datasets, namely CIFAR-10 and ImageNet running on networks well studied in the literature. Using one single MF-DFP network on our tastbenches, our design achieves up to 90\% energy savings with an insignificant accuracy drop of approximately 1\%. Using an ensemble of two networks, energy savings of 80\% are achievable while delivering accuracy gains of more than 1\% for CIFAR-10, and 0.5\% for ImageNet top-1 classification accuracy.

\section*{Acknowledgment}
This work is supported by NSF grant 1420864. We would like to thank NVIDIA Corporation for their generous GPU donation.

\setlength{\bibsep}{1pt plus 0.3ex}
\footnotesize
\bibliographystyle{abbrv}
\bibliographystyle{unsrtnat}
\scriptsize

\end{document}